\documentclass{article}

\PassOptionsToPackage{numbers, compress}{natbib}

 \usepackage[main, final]{neurips_2026}

\usepackage{amsmath,amsfonts,bm}

\def\eqref#1{equation~\ref{#1}}

\def\1{\bm{1}}

\DeclareMathAlphabet{\mathsfit}{\encodingdefault}{\sfdefault}{m}{sl}
\SetMathAlphabet{\mathsfit}{bold}{\encodingdefault}{\sfdefault}{bx}{n}

\usepackage{multirow}
\usepackage{color}
\usepackage[dvipsnames]{xcolor}
\usepackage{colortbl}

\definecolor{lavendergray}{rgb}{0.77, 0.76, 0.82}
\definecolor{lightgray}{rgb}{0.83, 0.83, 0.83}

\newcommand{\compare}[1]{{\cellcolor{Gray}#1}}

\usepackage{array}
\newcolumntype{H}{>{\setbox0=\hbox\bgroup}c<{\egroup}@{}}
\newcommand{\tablestyle}[2]{\setlength{\tabcolsep}{#1}\renewcommand{\arraystretch}{#2}\centering\small}

\usepackage{amssymb}
\usepackage{pifont}
\newcommand{\cmark}{\ding{51}}%
\newcommand{\xmark}{\ding{55}}%

\newcommand{\higherbetter}{$^\uparrow$}
\newcommand{\lowerbetter}{~$^\downarrow$}

\usepackage{wrapfig}

\definecolor{myblue}{rgb}{0.11764705882352941, 0.5647058823529412, 1.0}
\definecolor{Gray}{gray}{0.9}
\definecolor{darkgreen}{rgb}{0.545, 0.749, 0.608}
\definecolor{customblue}{rgb}{0.21,0.49,0.74}
\newcommand{\gain}[1]{{\color{darkgreen}\scriptsize\textbf{#1}}}

\usepackage{tablefootnote}

\usepackage{soul}
\sethlcolor{Gray}

\newif\ifdraft
\draftfalse

\usepackage[utf8]{inputenc} 
\usepackage[T1]{fontenc}    
\usepackage{url}            

\definecolor{myblue}{rgb}{0.21,0.49,0.74}
\usepackage[pagebackref,breaklinks,colorlinks,allcolors=myblue]{hyperref}

\hypersetup{
  colorlinks=true,
  urlcolor=myblue
}

\usepackage{booktabs}       
\usepackage{amsfonts}       
\usepackage{nicefrac}       
\usepackage{nicematrix}
\usepackage{microtype}      
\usepackage{multirow}
\usepackage{xcolor}         

\usepackage{arydshln}

\usepackage{xspace}
\newcommand{\ie}{\emph{i.e.}\xspace}

\newcommand{\eg}{\emph{e.g.}\xspace}

\usepackage{enumitem}
\usepackage[hypcap=false]{caption}

\usepackage{microtype}
\usepackage{graphicx}
\usepackage{subcaption}

\usepackage{amsmath}
\usepackage{amssymb}
\usepackage{mathtools}
\usepackage{amsthm}

\title{\underline{C}hannel-wise \underline{V}ector \underline{Q}uantization: Native 1D\\Visual Tokenization with Learnable Ordering}

\author{Wei Song$^{1,2,3}$\thanks{Co-first authors.}\enskip\enskip
    Tianhang Wang$^{1,3}$\footnotemark[1]\enskip\enskip
    Yitong Chen$^{1,4}$\enskip
    Tong Zhang$^{6}$\\
    \textbf{Zuxuan Wu$^{1,4}$\enskip
    Min Li$^{3}$\enskip
    Jiaqi Wang$^{1,5}$\thanks{Corresponding authors.}\enskip\thanks{Project lead.}\enskip\enskip
    Kaicheng Yu$^{2}$\footnotemark[2]
    }\\[.4ex]
    \normalsize \textsuperscript{1}Shanghai Innovation Institute\enskip
    \normalsize \textsuperscript{2}Westlake University\enskip 
    \normalsize \textsuperscript{3}Zhejiang University\enskip\\
    \normalsize \textsuperscript{4}Fudan University\enskip
    \normalsize \textsuperscript{5}JD.COM\enskip
    \normalsize \textsuperscript{6}University of Chinese Academy of Sciences
}

\begin{document}

\maketitle

\begin{figure*}[!ht]
  \centering
  \vspace{-1.1em}
  \includegraphics[width=0.975\textwidth]{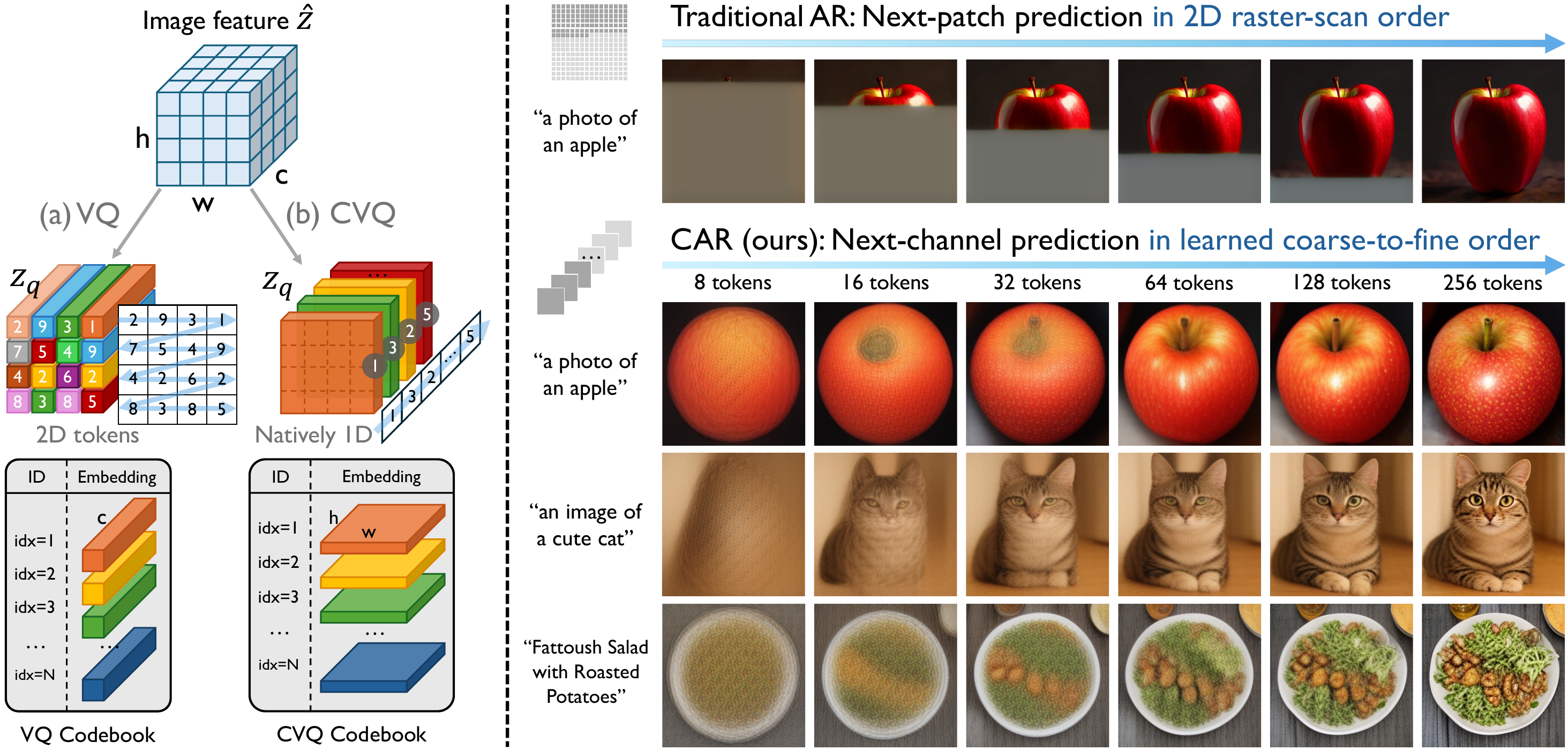}
  \caption{
      \textbf{(Left) VQ vs.\ CVQ.} Conventional VQ assigns an index to each $1\!\times\!1\!\times\!c$ patch feature vector, while CVQ assigns an index to each \textit{channel} of the feature map.
      \textbf{(Right) AR vs.\ CAR.} Traditional autoregressive (AR) models generate images patch by patch in raster scan order (here, Emu3~\citep{emu3}), whereas our channel-wise autoregressive (CAR) model generates images channel by channel in a 1D coarse-to-fine order.
      For example, given the prompt ``a photo of an apple'', the model first sketches a red circular shape corresponding to the apple's outline and dominant color, then progressively depicts its appearance, and finally adds fine-grained details such as specular highlights and yellow speckles.
  }
  \label{fig:visual_abstract}
\end{figure*}

\begin{abstract}

We present \textbf{\underline{C}hannel-wise \underline{V}ector \underline{Q}uantization (CVQ)}, a novel image tokenization paradigm that replaces patch-wise tokens with channel-wise tokens.
Unlike conventional vector quantization, which assigns a discrete token to each patch feature vector, CVQ quantizes each \textit{channel} of the feature map.
This represents an image as a 1D channel sequence with learnable coarse-to-fine ordering, rather than a 2D patch grid.
Based on CVQ, we introduce a new visual autoregressive framework with \textbf{\textit{``next-channel prediction''}}.
Instead of rendering images patch by patch in raster order, our \textbf{\underline{C}hannel-wise \underline{A}uto-\underline{R}egressive (CAR)} model predicts image channels sequentially, producing progressively enriched visual details. Specifically, it first sketches global structure and then refines fine-grained attributes, akin to a human artist's workflow.
Empirically, we show that:
(1) CVQ exhibits strong codebook scalability, maintaining 95\%+ utilization with up to $2^{16}$ entries while delivering growing reconstruction gains over VQ as the codebook scales up; and
(2) CAR attains a DPG score of 86.7 and a GenEval score of 0.79, demonstrating strong effectiveness for text-to-image generation.
The code is available \href{https://github.com/songweii/CVQ}{at this link}.

\end{abstract}

\section{Introduction}
\label{sec:intro}

Vector Quantization (VQ)~\citep{vq,esser2021vqgan} is a fundamental technique for discretizing continuous image representations and serves as a cornerstone of discrete image generation~\citep{chang2022maskgit,llamagen,chameleon}. However, since its introduction, the community has largely adhered to a patch-wise paradigm by default, where each index is assigned to represent a local $1\!\times\!1\!\times\!c$ feature vector, as illustrated in Fig.~\ref{fig:visual_abstract} (left).

We argue that this long-standing convention imposes two major limitations.
(1) \textit{Insufficient codebook usage}, which leads to severe information loss and, consequently, poor reconstruction quality. While prior efforts have attempted to mitigate this issue~\citep{vq-lc,simvq,rotationtrick}, these works typically rely on complex tricks or extra parameters that increase structural complexity, or require token factorization~\citep{vq-fc,llamagen}, which projects image features into a low-dimensional space for code index lookup. Such dimensionality reduction limits representational capacity~\citep{ibq} and substantially compromises token expressiveness~\citep{unitok}.
(2) \textit{Not naturally suited to sequence-to-sequence modeling}. Next-token prediction has achieved remarkable success in language models~\citep{gpt4,yang2025qwen3}, and the vision community seeks to mirror this success. However, language is inherently a 1D sequential signal (left-to-right), whereas images are spatial. Patch-wise tokenization discretizes images into 2D grids of tokens, which are then mechanically flattened into a 1D sequence to accommodate autoregressive (AR) learning (\eg, via raster scan or z-curve). This structural mismatch results in a suboptimal token ordering for unidirectional AR modeling~\citep{NFIG,RandAR,RAR}, as it disrupts local spatial dependencies among neighboring tokens~\citep{VAR,spectralar}.
Furthermore, as discussed in \citep{Hita}, the strong local spatial bias of patch tokens makes it difficult to impose an AR-friendly ordering on them (\eg, via nested dropout)~\citep{rippel2014learning}.

In this paper, we show that \textit{a simple change in the quantization axis} naturally resolves both limitations. Specifically, we shift VQ from a \textit{patch-wise} to a \textit{channel-wise} formulation. Unlike standard VQ, which learns a discrete \textit{spatial} codebook of patch-wise indices, our CVQ learns a discrete \textit{sequential} codebook of channel-wise indices.

\begin{minipage}[ht]{0.344\textwidth}
    Our motivation is intuitive: people typically draw by layering different levels of visual information to form a complete image.
    For example, when drawing an apple, an artist may first outline its overall shape and color tone before depicting finer details like the stem and speckles. Interestingly, as shown in Fig.~\ref{fig:feature_channels}, an autoencoder reflects a similar behavior by distributing different levels of visual information across channels, which jointly determine the final complete image.
\end{minipage}
\hfill
\begin{minipage}[ht]{0.64\textwidth}
    \centering
    \includegraphics[height=2.85cm]{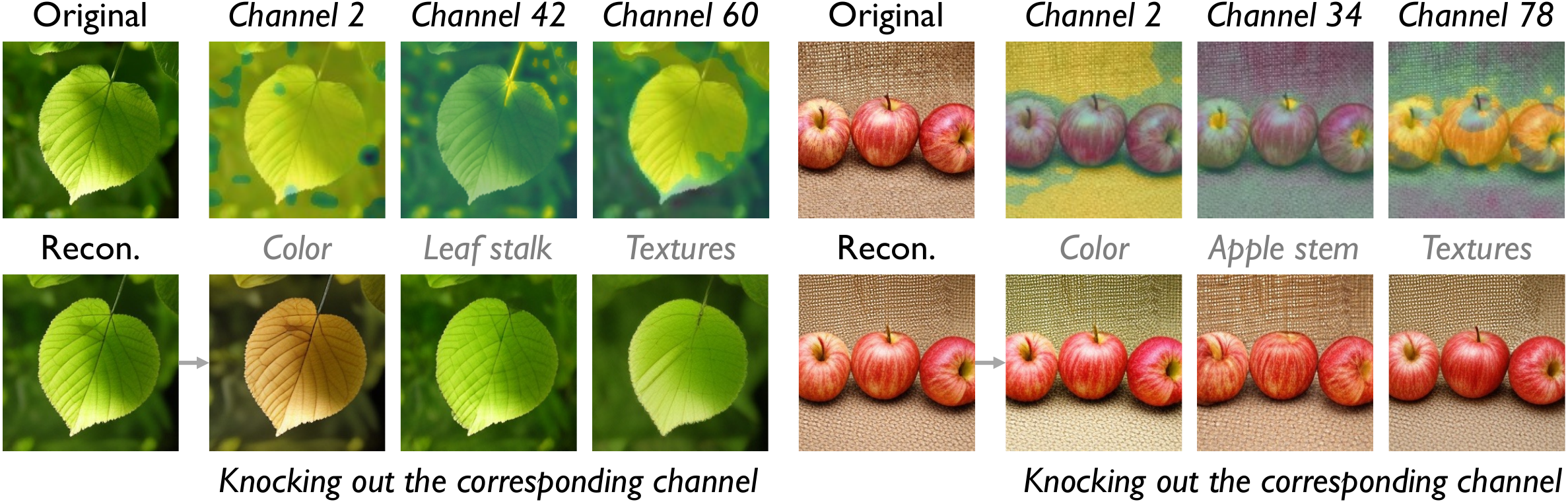}
    \captionsetup{type=figure}
    \caption{
        We visualize several channel activation maps from our VQ-VAE encoder and ablate individual channels by zeroing them before reconstruction. Interestingly, removing a channel selectively alters the corresponding content in the reconstructed image, with effects ranging from global appearance (\eg, leaf color) to fine-grained structure (\eg, removing the apple stem).
    }
    \label{fig:feature_channels}
\end{minipage}

Motivated by this, unlike conventional VQ, which discretizes spatial vectors at each x,y location, we propose to discretize each feature channel. Since each channel tends to capture different aspects of visual information, CVQ naturally represents an image as a 1D token sequence with progressively enriched visual content.
In contrast to conventional VQ, which imposes an artificial spatial ordering on the token sequence, CVQ internalizes spatial information within each token, as each channel encodes global spatial structure. As a result, the resulting token sequence forms a clean 1D representation, free from the limitations induced by patch-based tokenization.

On the other hand, as discussed in~\citep{vq-lc,simvq}, low codebook utilization in VQ arises from biased optimization dynamics, where only a small fraction of codebook vectors receive updates while the rest remain stagnant.
We argue that this issue stems from the redundancy among image patches~\citep{kaiming_patch}.
As analyzed in Fig.~\ref{fig:codebook} (a), for two images with similar textures, patch-wise partitioning produces highly overlapping embeddings, causing a large number of patch-wise embeddings within each training batch to be clustered to the same codebook indices (Fig.~\ref{fig:codebook} (b)). Consequently, codebook updates concentrate on only a small subset of entries, leading to dead codes and codebook collapse (Fig.~\ref{fig:codebook} (c)).
In contrast, partitioning features along the channel dimension yields more separable embeddings across images, resulting in broader codebook coverage and improved codebook utilization.

In summary, our contributions are as follows:
\begin{itemize}[leftmargin=*]
    \item We provide a new perspective on vector quantization, showing that a simple change in the quantization axis effectively mitigates codebook collapse without additional modules or constraints.
    \item We propose a novel visual tokenization paradigm that represents images as native 1D sequences of channels, rather than 2D grids of patches, and show that, with simple nested dropout~\citep{rippel2014learning}, the 1D channel tokens can learn a meaningful coarse-to-fine ordering that fits naturally with AR learning.
    \item Building on CVQ, we reformulate autoregressive image generation as a \textit{next-channel prediction} framework that progressively enriches visual content from global structures to fine-grained details.
\end{itemize}

\section{Related Work}
\label{related}

\textbf{Vector Quantization~}
Vector quantization is essential for discretizing visual signals into tokens.
VQ-VAE~\citep{vq} first introduced a learnable codebook to obtain discrete latent representations. Building upon this, VQGAN~\citep{esser2021vqgan} incorporates adversarial and perceptual losses to enhance image fidelity. RQ-VAE~\citep{lee2022rqvae} and MoVQ~\citep{zheng2022movq} further reduce quantization error through multi-stage quantization and vector modulation.
Despite these advances, low codebook utilization remains a central challenge in VQ. To mitigate this issue, ViT-VQGAN~\citep{vq-fc} proposes token factorization by projecting image features into a low-dimensional space for code index lookup. FSQ~\citep{fsq} and LFQ~\citep{magvit-v2} extend this idea by quantizing representations into a small set of fixed values to prevent codebook collapse. However, such approaches substantially limit representational capacity~\citep{unitok,ibq}.
Recent works, including VQGAN-LC~\citep{vq-lc} and SimVQ~\citep{simvq}, achieve higher codebook utilization through CLIP-feature initialization and learnable bases. However, these methods rely on more complex training pipelines and additional parameters compared to the original VQ architecture. IBQ~\citep{ibq} focuses on improving gradient propagation through index backpropagation and is orthogonal to our work.
To the best of our knowledge, \citep{anytime} is the only work that performs quantization along channels. However, it solely serves as an auxiliary component to support anytime sampling under computational constraints, and has not been systematically studied or evaluated as an independent vector quantization method.
So far, dominant VQ methods remain patch-wise and thus inherit the limitations of 2D grid tokenization.

\textbf{Autoregressive Visual Generation~}
Inspired by the success of Large Language Models (LLMs)~\citep{gpt3,gpt4}, autoregressive (AR) image generation has become a current research hotspot. Existing methods typically tokenize images into 2D grids using VQGAN-like models and flatten them into raster-scan sequences, creating a structurally misaligned ordering for next-token prediction~\citep{llamagen,chameleon,emu3}.
On the other hand, VAR~\citep{VAR} proposes next-scale prediction, which tokenizes images into multi-scale 2D tokens with bidirectional modeling within each scale. Infinity~\citep{infinity} further scales this approach to a much larger vocabulary size, achieving promising results. However, VAR deviates from the standard next-token prediction paradigm of LLMs, and its multi-scale hierarchy relies on heuristic partitioning.

\begin{minipage}[ht]{0.49\textwidth}
    Another line of work explores 1D visual tokenization. TiTok~\citep{titok}, SpectralAR~\citep{spectralar} and Hita~\citep{Hita} resample 2D image representations into 1D sequences via learnable queries, while FlexTok~\citep{bachmann2025flextok} and Semanticist~\citep{semanticist} extend the detokenizer to diffusion models. However, these methods rely on additional modules, such as token resampling modules and diffusion decoders~\citep{bachmann2025flextok,semanticist,catok}, leading to complex architectures, potential information bottlenecks, and increased learning difficulty.
    In contrast, CVQ operates at a different conceptual level. As illustrated in Fig.~\ref{fig:compare}, CVQ's 1D structure arises directly from the quantization process, enabling a natively 1D token sequence without specialized architectural design while maintaining the standard next-token prediction paradigm.
\end{minipage}
\hfill
\begin{minipage}[ht]{0.49\textwidth}
    \centering
    \includegraphics[height=5.35cm]{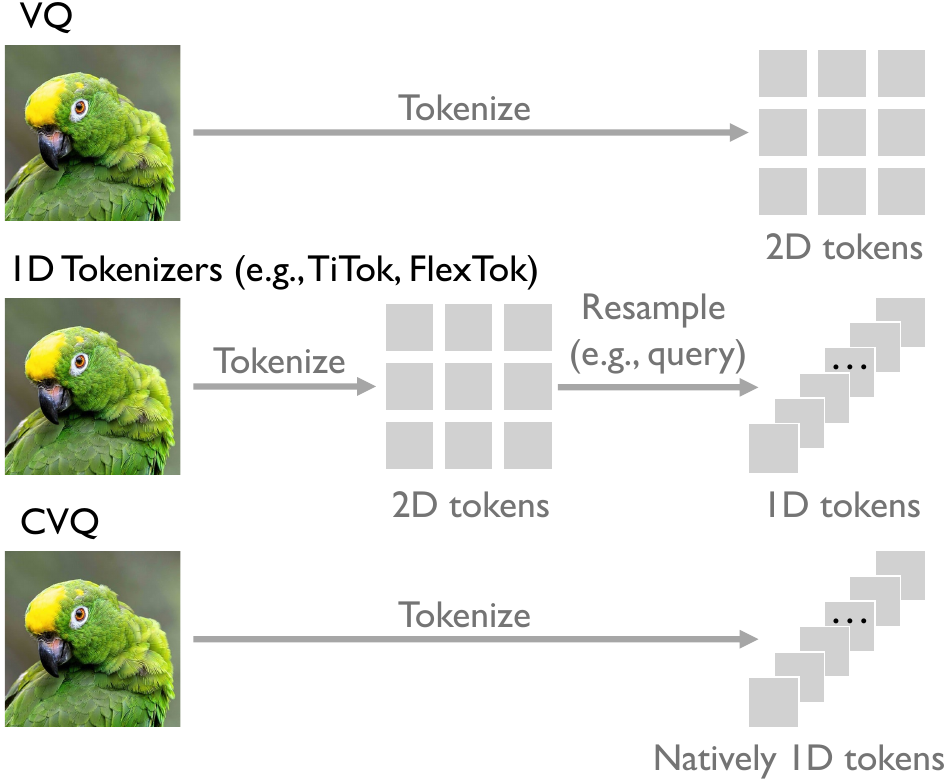}
    \captionsetup{type=figure}
    \caption{
        \textbf{Visual Tokenization Paradigms.}
    }
    \label{fig:compare}
\end{minipage}

\section{Method}
\label{method}

\begin{figure*}[t]
  \centering
  \includegraphics[width=\textwidth]{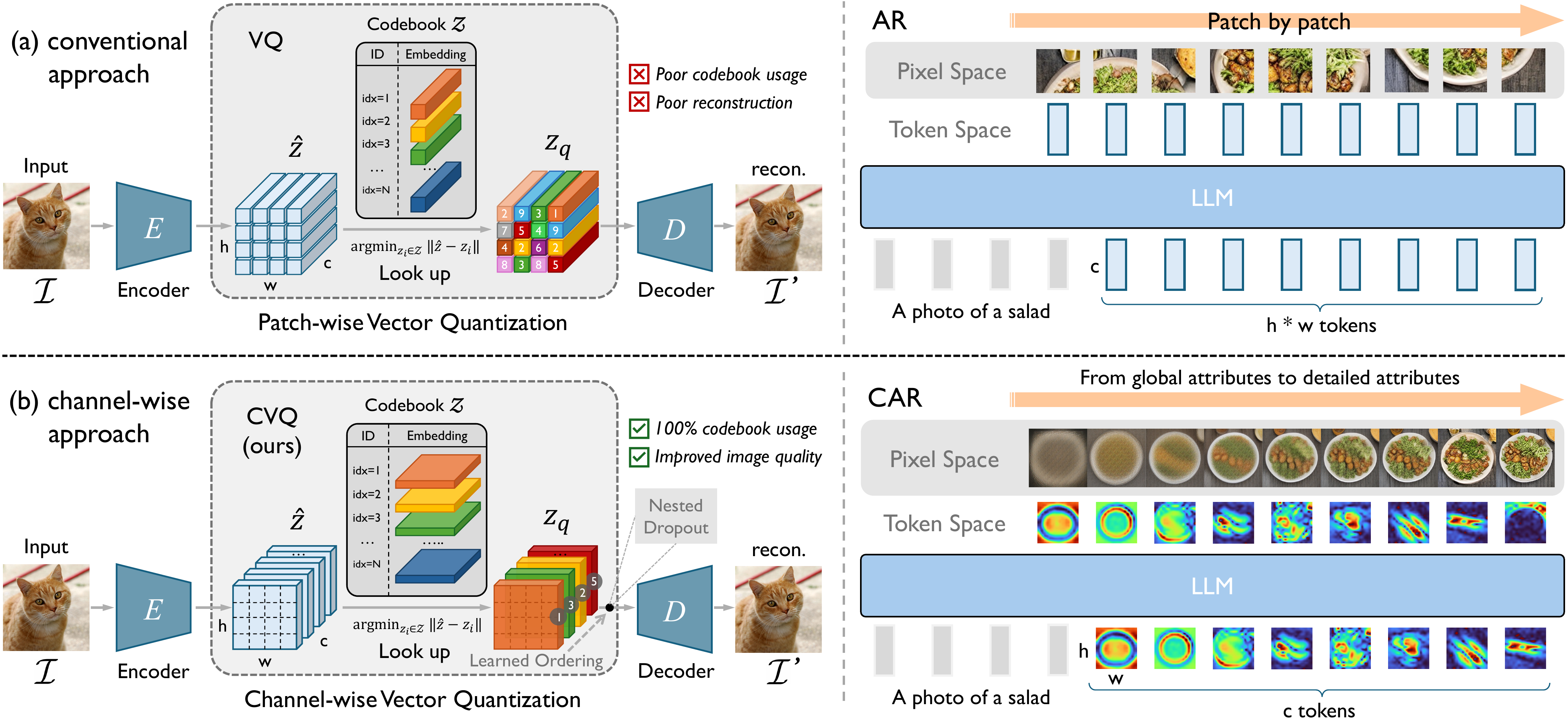}
  \caption{
      \textbf{(a) Conventional VQ and standard autoregressive image generation.} Conventional VQ assigns an index to each $1\!\times\!1\!\times\!c$ patch feature vector, which causes poor codebook usage and reconstruction quality. Furthermore, standard AR models predict tokens patch by patch, which imposes inherent sequential constraints on 2D spatial modeling.
      \textbf{(b) Proposed CVQ and CAR.} Our CVQ assigns an index to each $h\times w\times1$ channel feature vector, enabling high codebook utilization, improved reconstruction quality, and an orderable 1D token space. Building on this, the proposed CAR model predicts visual content hierarchically, progressing from global structures to fine details.
  }
  \label{fig:frameworks}
\end{figure*}

\subsection{Channel-wise Vector Quantization}

Traditional VQ frameworks map continuous latent variables to a discrete spatial codebook. As shown in Fig.~\ref{fig:frameworks}(a), given an input image $\mathcal{I}\in\mathbb{R}^{H \times W \times 3}$, it is encoded into a latent representation $\mathbf{Z} \in \mathbb{R}^{h \times w \times c}$, where $h=H/f$, $w=W/f$ and $f$ is the downsample ratio. In conventional VQ schemes, $\mathbf{Z}$ is treated as a collection of $M = h \times w$ spatial patch vectors, formulated as
\begin{equation}
    \mathbf{Z} =
\begin{bmatrix}
\mathbf{z}^{(1,1,:)} & \mathbf{z}^{(1,2,:)} & \cdots & \mathbf{z}^{(1,w,:)} \\
\mathbf{z}^{(2,1,:)} & \mathbf{z}^{(2,2,:)} & \cdots & \mathbf{z}^{(2,w,:)} \\
\vdots & \vdots & \ddots & \vdots \\
\mathbf{z}^{(h,1,:)} & \mathbf{z}^{(h,2,:)} & \cdots & \mathbf{z}^{(h,w,:)}
\end{bmatrix},
\end{equation}
where $\mathbf{z}^{(i,j,:)} \in \mathbb{R}^{1\times1\times c}$ denotes a spatial vector at the position $(i,j)$.
The quantization process involves a nearest-neighbor lookup within the codebook $\mathcal{C}_{\text{spatial}} = \{\mathbf{e}_n\}_{n=1}^N$, 
where $N$ denotes the codebook size:
\begin{equation}
    \mathbf{z}_{q}^{(i,j,:)} = \mathop{\mathrm{argmin}}_{\mathbf{e}_n \in \mathcal{C}_{\text{spatial}}} \left\| \mathbf{z}^{(i,j,:)} - \mathbf{e}_n \right\|_2^2. \label{eq:spatial}
\end{equation}
However, this conventional VQ paradigm typically suffers from codebook collapse, where only a small portion of the codebook receives gradient updates during training.~\citep{vq-lc,simvq,ibq}.

\begin{figure}[t]
  \centering
  \includegraphics[width=0.98\textwidth]{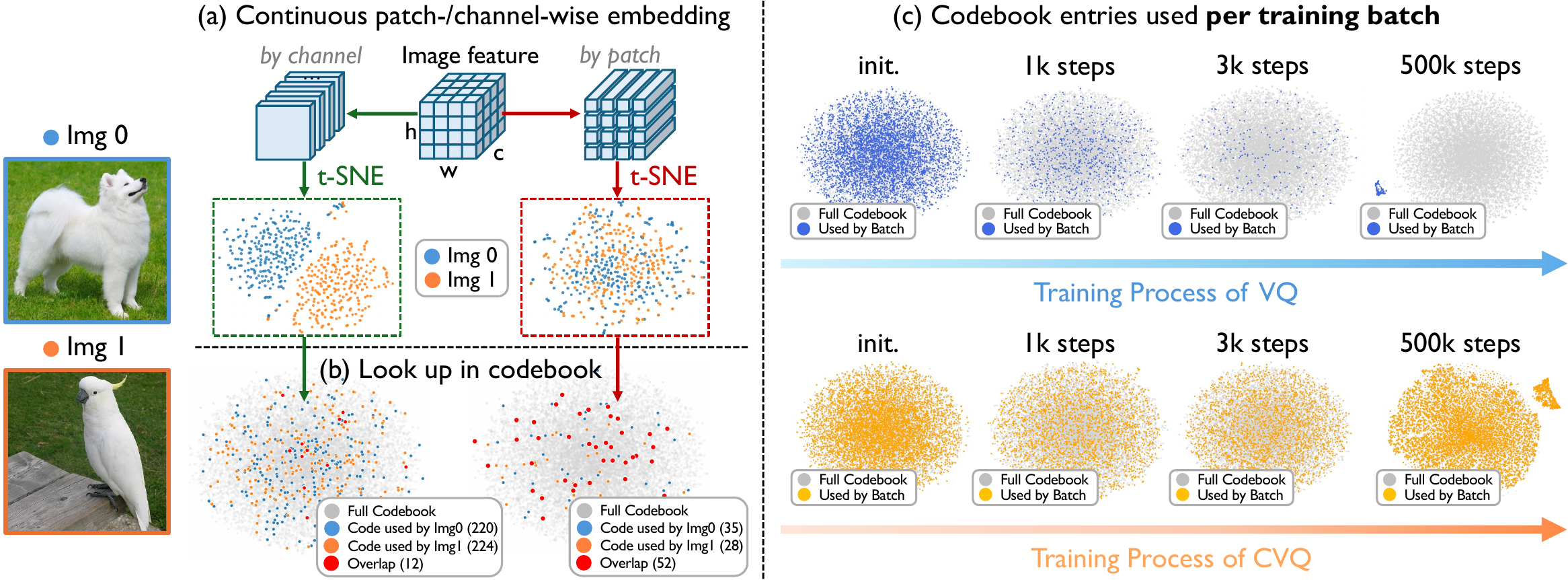}
  \caption{
    \textbf{(a) t-SNE visualization of embedding distributions.} For images with similar appearances but distinct semantics. Patch-wise partitioned embeddings of the two images are highly intermingled within a shared distribution, whereas channel-wise partitioning produces clearly separable clusters. 
    \textbf{(b) t-SNE of codebook vectors used by each image.} Patch similarity leads to significant intra- and inter-image overlap across codebook entries. In contrast, channel-wise embeddings are distributed over a broader range of codebook points, resulting in higher utilization. More results are provided in Appendix.~\ref{more_viz}.
    \textbf{(c) t-SNE of codebook entries used per training batch.} Conventional VQ rapidly collapses to a small subset of outlier entries, while CVQ maintains high codebook utilization throughout the training process.
  }
  \label{fig:codebook}
\end{figure}

We argue that this issue stems from the similarity and repetition between image patches~\citep{kaiming_patch}. 
As illustrated in Fig.~\ref{fig:codebook}(a), the patch-wise partitioned embeddings from different images are heavily entangled, resulting in substantial overlap of codebook indices both within and across images (Fig.~\ref{fig:codebook}(b)).
Such redundancy and recurrence cause patches to cluster around the same vectors during early training. This, in turn, leads to the ``death'' of other cluster centers, ultimately resulting in codebook collapse (Fig.~\ref{fig:codebook}(c)).

In contrast to conventional VQ, CVQ introduces a distinct quantization mechanism, as shown in Fig.~\ref{fig:frameworks}(b). The latent representation is partitioned as
\begin{equation}
    \mathbf{Z} = \bigl[
        \mathbf{z}^{(:,:,1)}, \mathbf{z}^{(:,:,2)}, \ldots, \mathbf{z}^{(:,:,c)}
    \bigr],
\end{equation}
where $\mathbf{z}^{(:,:,k)} \in \mathbb{R}^{h\times w\times 1}$ denotes the $k$th channel of $\mathbf{Z}$. 
For notational convenience, we denote $\mathbf{z}^{(:,:,k)}$ as $\mathbf{z}^{(k)}$ in the remainder of the paper. The quantization process is
\begin{equation}
    \mathbf{z}_{q}^{(k)} = \mathop{\mathrm{argmin}}_{\mathbf{e}_n \in \mathcal{C}_{\text{channel}}} \left\| \mathbf{z}^{(k)} - \mathbf{e}_n \right\|_2^2,
    \label{eq:channel}
\end{equation}
where $\mathcal{C}_{\text{channel}}$ represents the channel-wise codebook comprising codewords $\mathbf{e}_n \in \mathbb{R}^{h \times w \times 1}$. For the lookup of two-dimensional representations, we adopt a simple formulation based on the Frobenius norm, which is equivalent to flattening the matrix into a vector and performing a nearest-neighbor lookup. The forward pass is defined as
\begin{equation}
    \mathbf{z}_{q}^{(k)} = \mathbf{z}^{(k)} + \text{sg}[\mathbf{e}_n - \mathbf{z}^{(k)}],
\end{equation}
where $\text{sg}[\cdot]$ denotes the stop-gradient operator. For the backward pass, CVQ utilizes the standard \textit{Straight-Through Estimator} (STE)~\citep{vq} approach, where gradients are copied directly from the quantized representation $\mathbf{z}_{q}$ to the continuous latent $\mathbf{z}$.

Within the CVQ framework, each codebook index represents a global channel feature rather than a local spatial patch. As illustrated in Fig.~\ref{fig:codebook}(a), partitioning representations along the channel dimension yields features that are highly distinguishable across different images.
This property promotes high codebook utilization and reduces overlap, as evidenced in Fig.~\ref{fig:codebook}(b). Consequently, the t-SNE visualization in Fig.~\ref{fig:codebook}(c) shows that CVQ activates a significantly larger portion of the codebook within each training batch ($\text{batch size} = 64$) compared to standard VQ.

\textbf{Tokenizer Training}~~Following the standard VQGAN approach~\citep{esser2021vqgan}, the CVQ tokenizer is trained with a reconstruction objective comprising pixel-wise $\ell_2$ loss, commitment loss, LPIPS loss~\citep{zhang2018lpips}, and adversarial loss with PatchGAN discriminator~\citep{patchgan}.

\subsection{Channel-wise Autoregressive Generation}

As illustrated in Fig.~\ref{fig:frameworks}(b), we reformulate next-token prediction in autoregressive visual modeling by shifting from the traditional next-patch prediction paradigm to a \textit{Next-Channel Prediction} (NCP) strategy, where the autoregressive unit is a \textit{channel} rather than a patch token.

Through the CVQ tokenizer, an image is represented as a 1D sequence of $c$ discrete channel tokens, $\mathcal{X} = \{x_i\}_{i=1}^{c}$. Subsequently, the decoder-only transformer is trained autoregressively to predict the next channel token conditioned on the textual context. The autoregressive likelihood of the entire image is thus given by
\begin{equation}
p(\mathcal{X}) = \prod_{k=1}^{c} p\!\left(x^{(k)} \mid x^{(<k)}\right).
\end{equation}
To feed the channel tokens with dimension $h \times w$ into the transformer backbone, a two-layer MLP projector is applied to align their dimensionality with the LLM backbone. 

\textbf{Inducing Prefix-Based Channel Ordering}~~Since channels do not possess an inherent order, we apply \textit{nested channel dropout} during tokenizer training to induce a meaningful coarse-to-fine ordering for autoregressive modeling. Specifically, given a sequence of $c$ channel tokens, we randomly sample a prefix length $c_{\text{keep}}\in\{1,\ldots,c\}$ and retain only the first $c_{\text{keep}}$ channels, masking the remaining channels to zero.
By enforcing reconstruction from variable-length channel prefixes, nested dropout induces a \textit{coarse-to-fine} hierarchy, where early channels capture global structure and later channels progressively encode finer details~\citep{rippel2014learning}. 
This learned ordering enables progressive reconstruction as more channels are retained, as visualized in Fig.~\ref{fig:progressive}. We further analyze the effect of nested channel dropout in Sec.~\ref{sec:exp_discussion}, with additional details provided in Appendix~\ref{appendix_channel_dropout}.
\section{Experiments}
\label{experiment}

This section provides a comprehensive experimental analysis of CVQ's performance in downstream image reconstruction and generation tasks. In the controlled 256-token comparison, we set $c = h \times w = 256$, so CVQ and patch-wise VQ quantize the same latent tensor with identical token counts, codeword dimensions, and nearest-neighbor lookup complexity.

\subsection{Visual Reconstruction}
\label{sec:exp_recon}

\begin{table}[t]
    \centering
    \captionsetup{type=table}
    \caption{
        \textbf{Reconstruction performance of different VQ methods on ImageNet-1K.} The vanilla form of CVQ improves reconstruction fidelity over VQ-based methods across both 256- and 1024-token budgets, outperforming established baselines such as MoVQ and VQ-LC at comparable codebook scales. Following prior work~\citep{vq-lc,simvq,ibq}, we define codebook usage as the percentage of codebook entries selected at least once when tokenizing all 50,000 images in the ImageNet-1K validation set. $\dagger$ denotes models we reproduced using the same training setup as our CVQ tokenizer.
    }
    \label{tab:recon}
    \tablestyle{2pt}{1.108}
    \resizebox{0.8\linewidth}{!}{
    \begin{tabular}{lccccccc}
    \toprule
    \multirow{2}{*}{\textsc{VQ Methods}}  &\multirow{2}{*}{\# Tokens} &\multirow{2}{*}{Ratio} &\multirow{2}{*}{\shortstack{Codebook\\Size}} &\multirow{2}{*}{rFID\lowerbetter} &\multirow{2}{*}{SSIM\higherbetter} &\multirow{2}{*}{PSNR\higherbetter} &\multirow{2}{*}{\shortstack{Codebook\\Usage\higherbetter}} \\
    \\
    \midrule
    \multicolumn{6}{l}{\textbf{\textit{256 tokens}}}\\
    Vanilla VQ$^{\dagger}$~\citep{esser2021vqgan} &16$\times$16  &16            &16,384           &4.84                    &0.542                       &19.93           &4.5\%\\
    Dynamic VQ~\citep{DynamicVQ}                  &16$\times$16  &\{8, 16, 32\} &1,024            &4.08                    &-                           &-               &-\\
    Dynamic VQ$^{\dagger}$~\citep{DynamicVQ}      &16$\times$16  &\{8, 16, 32\} &16,384           &3.62                    &0.558                       &20.07           &5.2\%\\
    VQ-LC~\citep{vq-lc}                           &16$\times$16  &16            &16,384           &3.01                    &0.564                       &-               &99.0\%\\
    VQ-LC~\citep{vq-lc}                           &16$\times$16  &16            &100,000          &\underline{2.62}        &\textbf{0.589}              &-               &99.0\%\\
    SimVQ$^{\dagger}$~\citep{simvq}               &16$\times$16  &16            &16,384           &2.63                    &0.556                       &\textbf{21.10}  &\textbf{100.0\%}\\
    \compare{}Vanilla CVQ (ours)                  &\compare{}256 &\compare{}16  &\compare{}16,384 &\compare{}\textbf{2.60} &\compare{}\underline{0.565} &\compare{}\underline{20.94} &\compare{}\textbf{100.0\%}\\
    \midrule
    \multicolumn{6}{l}{\textbf{\textit{1024 tokens}}}\\
    Vanilla VQ$^{\dagger}$~\citep{esser2021vqgan} &32$\times$32          &8            &16,384           &1.32                    &0.692                    &23.35       &2.8\%\\
    VQ w/ rotation trick~\citep{rotationtrick}    &32$\times$32          &8            &16,384           &1.10                    &-                        &-           &27.0\%\\
    VQ-FC~\citep{vq-fc}                           &32$\times$32          &8            &8,192            &1.28                    &-                        &-           &95.0\%\\
    VQ-LC~\citep{vq-lc}                           &32$\times$32          &8            &100,000          &1.29                    &\underline{0.716}        &-           &99.5\%\\
    RVQ~\citep{lee2022rqvae}                      &8$\times$8$\times$16  &32           &16,384           &1.83                    &-                        &-           &-\\
    MoVQ~\citep{zheng2022movq}                    &16$\times$16$\times$4 &16           &1,024            &1.12                    &0.673                    &22.42       &-\\
    MoVQ~\citep{zheng2022movq}                    &16$\times$16$\times$4 &16           &16,384           &1.05                    &-                        &-          &-\\
    SimVQ$^{\dagger}$~\citep{simvq}               &32$\times$32          &8            &16,384           &\underline{0.90}        &0.714                    &\underline{24.88}  &\textbf{100.0\%}\\
    \compare{}Vanilla CVQ (ours)                  &\compare{}1024        &\compare{}16 &\compare{}16,384 &\compare{}\textbf{0.88} &\compare{}\textbf{0.723} &\compare{}\textbf{25.02} &\compare{}\textbf{100.0\%}\\
    \bottomrule
    \end{tabular}}
\end{table}

\textbf{Training Setup~~}
We train two versions of our vision tokenizer with $c=256$ and $c=1024$ latent channels, yielding 256 / 1024 channel tokens, respectively. Both use a $16\times16$ latent grid, so each codeword has 256 dimensions. Unless otherwise specified, we use a codebook size of 16,384. All models are trained on ImageNet-1K~\citep{deng2009imagenet} at $256\times256$ resolution for 100 epochs. We use Adam ($\beta_1=0.5$, $\beta_2=0.9$) with a learning rate of $1\times10^{-4}$, weight decay of $10^{-4}$, and a global batch size of 256. We further extend CVQ to variable resolution in Appendix~\ref{appendix_variable_resolution}.

\textbf{Main Results~~}
We measured reconstruction FID (rFID), PSNR, and SSIM on the ImageNet-1K (val).
As shown in Table~\ref{tab:recon}, compared to conventional VQ, whose codebook utilization collapses to only 4.5\% when scaling the codebook size to 16,384, CVQ natively supports 100\% codebook utilization without requiring any additional modifications.
This outperforms prior VQ improvement methods such as VQGAN-FC, which require additional token factorization, and VQGAN-LC, which relies on initializing the codebook using features extracted from a pretrained CLIP model and introduces extra projector parameters. Consequently, CVQ demonstrates a remarkable and consistent improvement over traditional VQ-based methods across different token budgets.
With 256 tokens, CVQ achieves an rFID of 2.60, significantly improving over vanilla VQGAN (4.84) and outperforming SimVQ (2.63) in reconstruction fidelity. Under the 1024-token setting, the improvement becomes more pronounced: CVQ attains a lower rFID of 0.88 and a higher PSNR of 25.02 dB, surpassing strong baselines such as MoVQGAN (1.05 rFID) and VQGAN-LC (1.29 rFID).

\subsection{Visual Generation}
\label{sec:exp_gen}

\newcommand{\tstrut}{\rule{0pt}{11pt}}
\newcommand{\bstrut}{\rule[-1pt]{0pt}{1pt}}

\begin{table}[t]
\centering
\caption{\textbf{Quantitative comparison on text-to-image generation benchmarks.} We categorize methods by attention mask into \textit{Bidirectional}, \textit{Unidirectional}, and \textit{Hybrid} approaches. Bidirectional methods include diffusion and MaskGIT; unidirectional methods follow the standard AR paradigm of next-token prediction; while hybrid methods include the VAR family, which uses block-wise attention masks with full attention within each block. \textit{NPP}, \textit{NSP}, and \textit{NCP} denote \textit{next-patch}, \textit{next-scale}, and \textit{next-channel prediction}, respectively. Among them, only NPP and NCP follow the standard next-token prediction paradigm. $\dagger$ indicates results obtained with prompt rewriting or self-CoT.}
\label{tab:gen}
\resizebox{\linewidth}{!}{
\begin{tabular}{llcccccccc}
    \toprule
    \multirow{2}{*}{Type}&\multirow{2}{*}{Methods} &\multirow{2}{*}{\# Para.} &\multicolumn{4}{c}{GenEval$\uparrow$} &\multicolumn{3}{c}{DPG$\uparrow$} \\
    \cmidrule(l){4-7} \cmidrule(l){8-10} & & & Two Obj. & Position & Color Attri. & \textbf{Overall} & Global & Relation & \textbf{Overall} \\
    
    \midrule
    \multicolumn{10}{c}{\textit{Bidirectional}} \\
    \midrule
    
    \multirow{6}{*}{\textit{Diff.}}
    &SDXL~\citep{sdxl}                 &2.6B  &0.74   &0.15  &0.23  &0.55  &83.27 &86.76 &74.65 \\
    &SD3 (d=38)~\citep{SD3}            &8B    &0.89   &0.34  &0.47  &0.71  &-     &-     &-     \\
    &Lumina-Next~\citep{lumina-next}   &1.7B  &0.49   &-     &0.15  &0.55  &86.89 &86.59 &80.50 \\
    &FLUX-dev~\citep{FLUX}             &12B   &-      &-     &-     &0.67  &-     &-     &84.00 \\
    &FLUX-schnell~\citep{FLUX}         &12B   &-      &-     &-     &0.71  &-     &-     &84.80 \\
    &SANA~\citep{sana}                 &1.6B  &0.77   &-     &0.47  &0.66  &-     &91.90 &84.80 \\
    
    \noalign{\vspace{1.5pt}}\hdashline\noalign{\vspace{1.5pt}}
    
    \multirow{2}{*}{\textit{Mask.}}
    &TiTok~\citep{titok}               &0.6B  &-      &-     &-     &0.49  &-     &-     &- \\
    &TA-TiTok~\citep{tatitok}          &1.1B  &0.58   &0.13  &0.34  &0.55  &-     &-     &- \\

    \midrule
    \multicolumn{10}{c}{\textit{Hybrid}} \\
    \midrule
    
    \multirow{8}{*}{\textit{NSP$^*$}}
    &VARGPT-1.1~\citep{vargpt1d1}      &9B           &0.53             &0.13             &0.21             &0.53             &84.83           &88.13           &78.59 \\
    &STAR~\citep{STAR}                 &1.7B         &-                &-                &-                &0.55             &-               &-               &-     \\
    &TokenFLow~\citep{tokenflow}       &7B           &0.72             &0.45             &0.42             &0.63             &78.72           &85.22           &73.38 \\
    &HART~\citep{hart}                 &0.7B         &-                &-                &-                &0.56             &-               &-               &80.89 \\
    &SWITTI~\citep{switti}             &3B           &-                &-                &-                &0.62             &-               &-               &-     \\
    &UniGen-AR~\citep{unigenar}        &2B           &0.76             &0.41             &0.45             &0.68             &-               &-               &- \\
    &Infinity~\citep{infinity}         &2B           &0.85$^{\dagger}$ &0.49$^{\dagger}$ &0.57$^{\dagger}$ &0.73$^{\dagger}$ &93.11           &90.76           &83.46 \\
    &InfinityStar~\citep{infinitystar} &8B           &0.90$^{\dagger}$ &0.62$^{\dagger}$ &0.67$^{\dagger}$ &0.79$^{\dagger}$ &91.68           &91.87           &86.55 \\
    
    \midrule
    \multicolumn{10}{c}{\textit{Unidirectional}} \\
    \midrule
    
    \multirow{10}{*}{\textit{NPP$^*$}}
    &LlamaGen~\citep{llamagen}        &0.8B          &0.34             &0.07             &0.04             &0.32             &-               &-               &64.84 \\
    &Chameleon~\citep{chameleon}      &7B            &-                &-                &-                &0.39             &-               &-               &- \\
    &Emu3~\citep{emu3}                &8B            &0.81$^{\dagger}$ &0.49$^{\dagger}$ &0.45$^{\dagger}$ &0.66$^{\dagger}$ &-               &-               &80.60 \\
    &Lumina-mGPT~\citep{lumina-next}  &7B            &0.77             &-                &0.32             &0.56             &-               &91.29           &79.68 \\
    &Janus~\citep{wu2024janus}        &1.5B          &0.68             &0.46             &0.42             &0.61             &82.33           &85.46           &79.68 \\
    &SimpleAR~\citep{simplear}        &1.5B          &0.90             &0.28             &0.45             &0.63             &87.97           &88.33           &81.97 \\
    &Liquid~\citep{liquid}            &7B            &0.73             &0.17             &0.37             &0.55             &-               &-               &- \\
    &UniTok~\citep{unitok}            &7B            &0.71             &0.26             &0.45             &0.59             &-               &-               &- \\
    &MUSE-VL~\citep{musevl}           &7B            &-                &-                &-                &0.57$^{\dagger}$ &-               &-               &- \\
    &NextStep-1~\citep{nextstep1}     &14B           &-                &-                &-                &0.73$^{\dagger}$ &-               &-               &85.28 \\
    
    \hdashline
    
    \multirow{2}{*}{\textit{NCP$^*$}}
    &\compare{}CAR (4B)\tstrut        &\compare{}4B  &\compare{}0.86   &\compare{}0.62   &\compare{}0.59   &\compare{}0.74   &\compare{}86.93 &\compare{}93.77 &\compare{}82.14 \\
    &\compare{}CAR (8B)               &\compare{}8B  &\compare{}0.92   &\compare{}0.66   &\compare{}0.66   &\compare{}0.79   &\compare{}89.40 &\compare{}94.16 &\compare{}86.72 \\
    \bottomrule
    \end{tabular}
}
\end{table}

\textbf{Training Recipe~~} 
Our CAR models are initialized from the pre-trained Qwen3-4B/8B backbones \citep{yang2025qwen3}. The training process is conducted on 80M text-image pairs and is divided into two stages:

\begin{itemize}[leftmargin=*]
    \item \textbf{Stage I:} To align the CVQ features with the LLM latent space, we employ a 2-layer MLP projector. This projector maps the 256-dimensional channel embeddings to the hidden dimension of the LLM backbones (2560 for the 4B model and 4096 for the 8B model). During this stage, only the MLP projector and the LLM head are optimized, while the LLM backbone remains frozen.
    \item \textbf{Stage II:} In the second stage, we perform end-to-end optimization across all parameters, including the MLP projector, the LLM backbone, and the LLM head. We list the data sources and training hyperparameters in the Appendix.
\end{itemize}

\begin{figure*}[t]
  \centering
  \includegraphics[width=\textwidth]{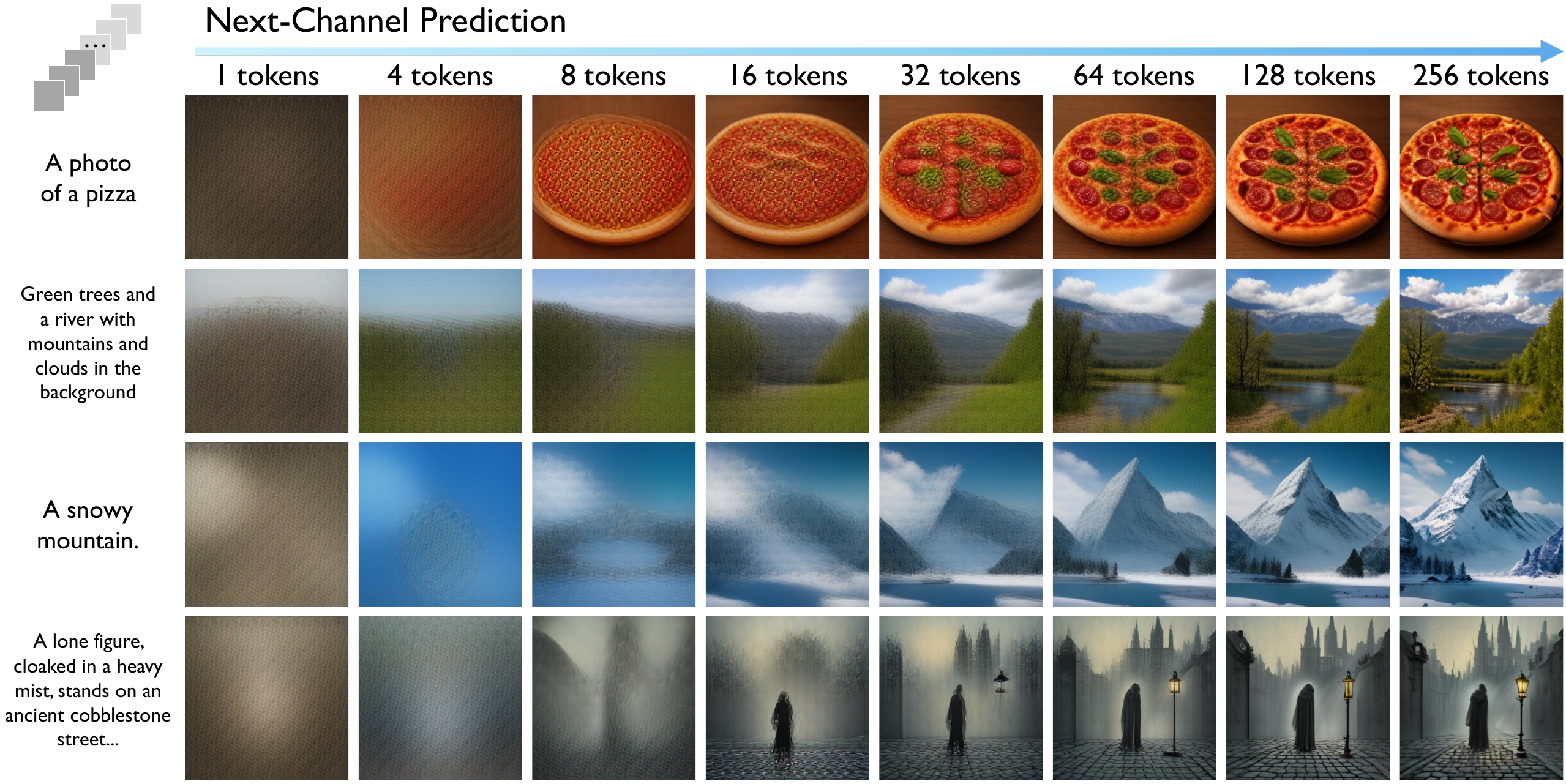}
  \caption{
      \textbf{The generation process of CAR.}
  }
  \label{fig:progressive}
\end{figure*}

\begin{figure*}[t]
  \centering
  \includegraphics[width=0.999\textwidth]{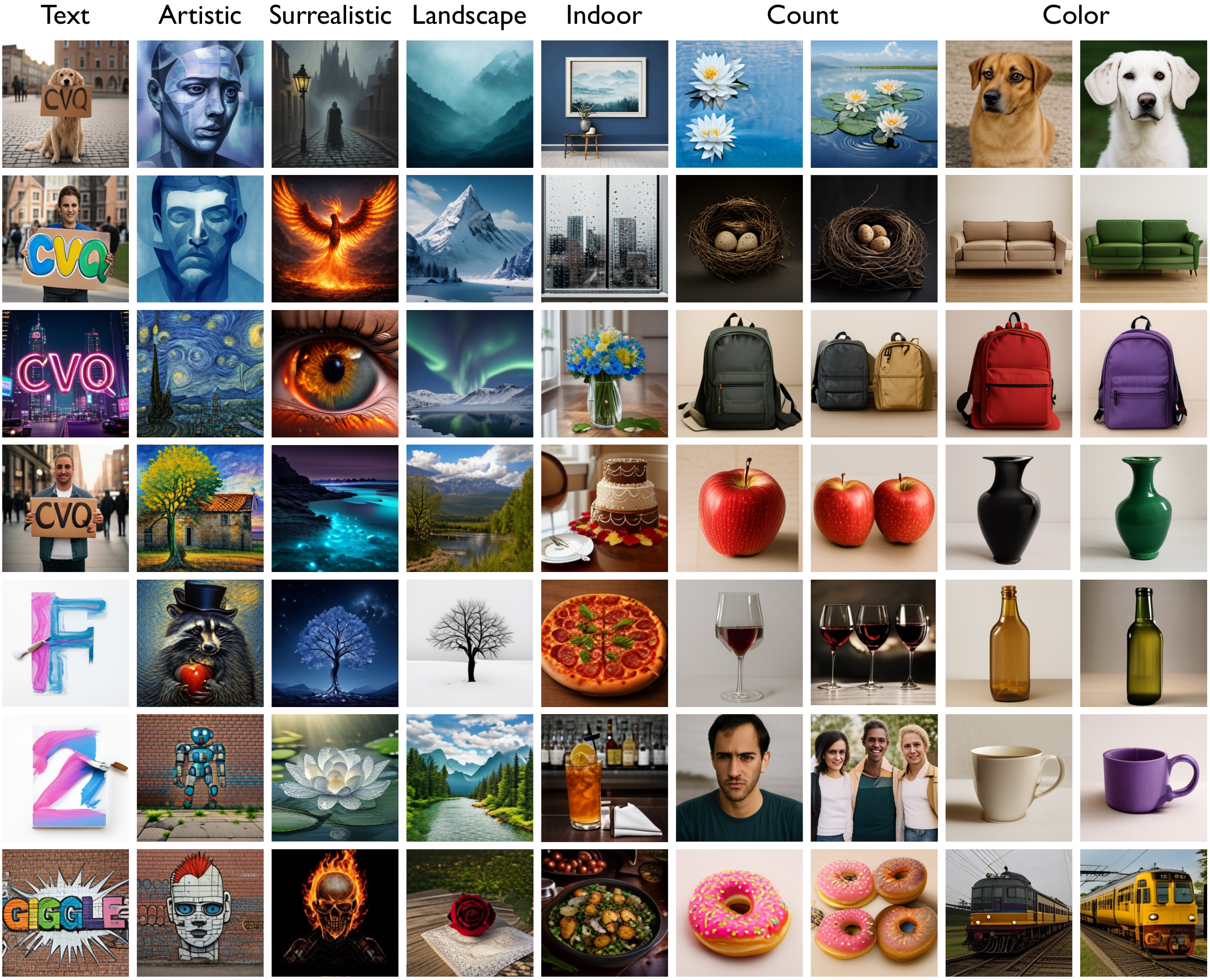}
  \caption{\textbf{Qualitative results on text-to-image generation.} 
  }
  \vspace{-0.1cm}
  \label{fig:gen_cases}
\end{figure*}

\textbf{Main Results~~}
As illustrated in Fig.~\ref{fig:progressive}, CAR produces progressively detailed image content as more channels are generated.
As shown in Table~\ref{tab:gen}, among unidirectional methods, CAR (4B) achieves competitive or superior performance compared with strong AR baselines such as NextStep-1 (14B) and Emu3 (8B). Scaling CAR to 8B further improves performance, reaching a GenEval score of 0.79 and a DPG overall score of 86.72, competitive with strong VAR methods such as Infinity and InfinityStar. These results suggest that CVQ provides an effective token ordering for AR learning while preserving a simple next-token prediction formulation. Qualitative results are shown in Fig.~\ref{fig:gen_cases}.

In addition to semantic-alignment benchmarks, we report MJHQ-30K FID in Table~\ref{tab:mjhq}. CAR achieves an FID of 6.42, outperforming both 1D masked-token baselines and the standard 2D-token baseline.

\subsection{Discussions}
\label{sec:exp_discussion}

\begin{minipage}[ht]{0.49\textwidth}
    \textbf{Codebook Scalability}~~As shown in Table~\ref{tab:codebook_utilization}, scaling up the codebook size leads to a severe decline in utilization for patch-wise VQ, dropping from 20.3\% to 1.1\%, with only marginal improvements in rFID. In contrast, CVQ maintains nearly 100\% utilization, even at massive scales (up to $65,536$). This enables CVQ to effectively leverage larger vocabularies for improved performance, reducing rFID from 3.64 to 2.32. Notably, such advantage becomes increasingly evident at scale: at a codebook size of 65K, CVQ achieves a 52\% improvement in reconstruction fidelity over the VQ baseline, exhibiting promising scaling properties.
\end{minipage}
\hfill
\begin{minipage}[ht]{0.49\textwidth}
    \centering
    \includegraphics[height=2.44cm]{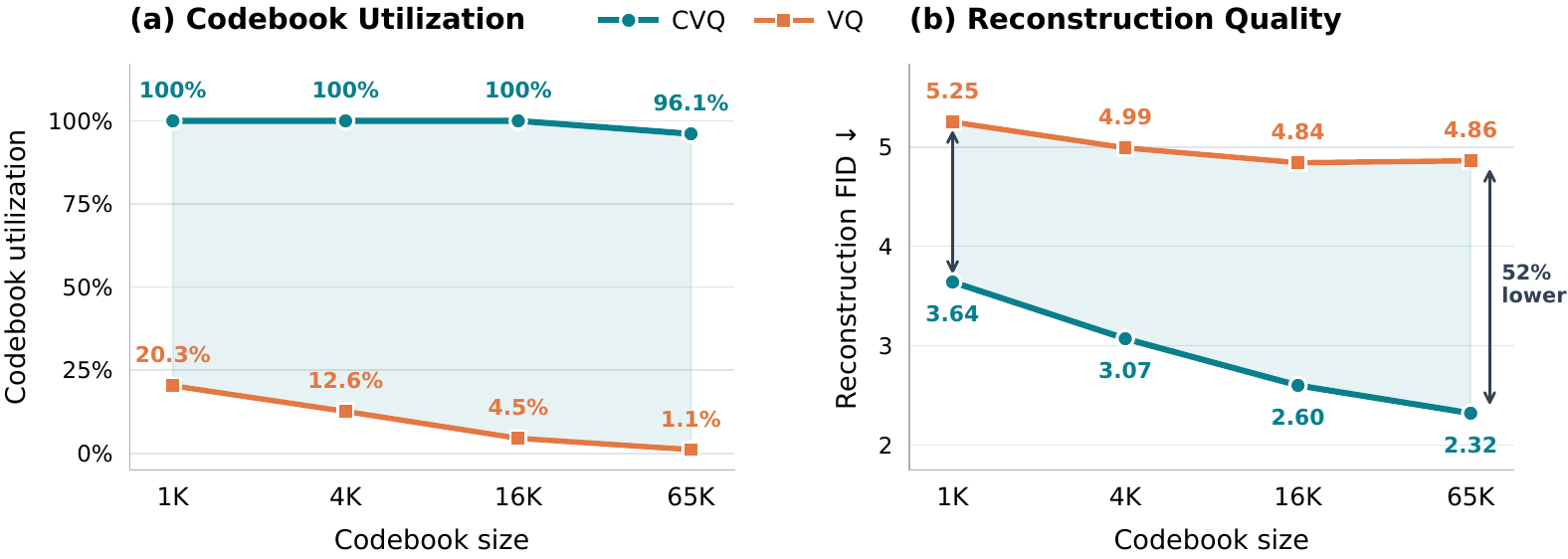}
    \captionsetup{type=figure}
    \caption{
        \textbf{Codebook Scalability.} (Left) CVQ maintains over 95\% utilization even with $2^{16}$ entries, while VQ suffers from codebook collapse as the codebook size increases. (Right) CVQ consistently improves reconstruction quality as the codebook scales up, with growing gains over VQ.
    }
    \label{fig:codebook_scale}
\end{minipage}

\textbf{Nested Dropout}~~As shown in Table~\ref{tab:w_wo_dropout}, CVQ without nested dropout exhibits weaker AR generation performance, as channel tokens do not possess an inherent order, whereas patch tokens still retain a natural spatial ordering despite suboptimal raster scanning. Yet, compared with 2D patch tokens, the 1D nature of channel-wise tokens makes it easy to impose a meaningful AR-friendly ordering.
With a simple nested dropout strategy~\citep{rippel2014learning}, CVQ learns a coarse-to-fine channel order, effectively improving AR performance while maintaining reconstruction quality. In contrast, such an ordering strategy is difficult to apply to 2D patch tokenization, as each patch token is strongly tied to local content and therefore cannot naturally represent a global coarse-to-fine progression.

\textbf{Versus 1D Tokenizers~~}
While related in target, CVQ and existing 1D tokenizer works operate at different conceptual levels: the former is a quantization method, whereas the latter primarily focus on additional modifications at the model architecture level (\eg, via learnable queries~\citep{titok,tatitok,spectralar} or diffusion decoders~\citep{bachmann2025flextok,semanticist,flowtok}). CVQ's 1D structure arises directly from the quantization process itself, enabling a 1D token sequence without specialized model architectural design.

Here, we provide a direct comparison with recent 1D tokenizers. Since these methods typically employ stronger training recipes, such as two-stage training, external proxy codes, or enhanced networks for LPIPS and GAN losses, we report results under both the standard VQGAN-style training protocol and a stronger TA-TiTok-style recipe in Table~\ref{tab:1d_recon}.

\vspace{0.1cm}

\belowrulesep=0pt\aboverulesep=0pt

\begin{minipage}[ht]{0.49\textwidth}
    \centering
    \captionsetup{type=table}
    \caption{
      \textbf{Ablation on codebook usage across different codebook scales.} Conventional VQ suffers from increasingly severe codebook collapse as the vocabulary grows, whereas CVQ maintains high utilization even with large codebooks, demonstrating superior codebook scalability.
    }
    \label{tab:codebook_utilization}
    \tablestyle{4pt}{1.18}
    \resizebox{\linewidth}{!}{
    \begin{tabular}{l|cc|cc}
    \toprule
    \multirow{2}{*}{Codebook Size} &\multicolumn{2}{c}{VQ}                &\multicolumn{2}{c}{CVQ} \\ 
    \cmidrule{2-5}                 &Util.\higherbetter &rFID\lowerbetter  &Util.\higherbetter    &rFID\lowerbetter \\
    \midrule
     1,024                         &20.3\%             &5.25              &100\%~\gain{(+79.7)}  &3.64\gain{($^\downarrow$31\%)} \\
     4,096                         &12.6\%             &4.99              &100\%~\gain{(+87.4)}  &3.07\gain{($^\downarrow$38\%)} \\
     16,384                        &4.5\%              &4.84              &100\%~\gain{(+95.5)}  &2.60\gain{($^\downarrow$46\%)} \\
     65,536                        &1.1\%              &4.86              &96.1\%~\gain{(+95.0)} &2.32\gain{($^\downarrow$52\%)} \\
    \bottomrule
    \end{tabular}}
\end{minipage}
\hfill
\begin{minipage}[ht]{0.49\textwidth}
    \centering
    \captionsetup{type=table}
    \caption{
        \textbf{Ablation on nested dropout.} (1) Standard patch-wise VQ is incompatible with nested dropout; (2) nested channel dropout effectively induces an AR-friendly ordering over channel tokens, substantially improving generation performance (GenEval +0.12, DPG +9.38) without compromising reconstruction quality.
    }
    \vspace{-0.05cm}
    \label{tab:w_wo_dropout}
    \tablestyle{4pt}{1.18}
    \resizebox{\linewidth}{!}{
    \begin{tabular}{l|ccc|cc}
    \toprule
    \multirow{2}{*}{Method} &\multicolumn{3}{c}{Recon.}                               &\multicolumn{2}{c}{Generation} \\ 
    \cmidrule{2-6}          &rFID\lowerbetter &SSIM\higherbetter  &PSNR\higherbetter  &GenEval\higherbetter  &DPG\higherbetter \\
    \midrule
     VQ w/o dropout       &4.84             &0.542              &19.93              &0.69                  &79.86 \\
     VQ w/ dropout        &15.32            &0.322              &3.43               &-                     &- \\
     \midrule
     CVQ w/o dropout      &\textbf{2.60}             &0.565              &20.94              &0.62                  &72.76 \\
     CVQ w/ dropout       &2.62             &\textbf{0.568}              &\textbf{21.01}              &\textbf{0.74}~\gain{(+0.12)}   &\textbf{82.14}~\gain{(+9.38)} \\
    \bottomrule
    \end{tabular}}
\end{minipage}

\begin{minipage}[ht]{0.49\textwidth}
    \centering
    \captionsetup{type=table}
    \caption{
        \textbf{Comparison with representative 1D tokenizers.} $\dagger$: standard VQGAN-style recipe, $\ddagger$: two-stage recipe with decoder reinforcement~\citep{titok}, $*$: models reproduced with the official codebase.
    }
    \label{tab:1d_recon}
    \tablestyle{3pt}{1.05}
    \resizebox{\linewidth}{!}{
    \begin{tabular}{lcccc}
    \toprule
    Method & Diff. Dec. & rFID\lowerbetter & SSIM\higherbetter & PSNR\higherbetter \\
    \midrule
    FlexTok~\citep{bachmann2025flextok}       &\cmark           &4.20                    &-                        &- \\
    FlexTok-1.4B~\citep{bachmann2025flextok}  &\cmark           &1.45                    &0.465                    &18.53 \\
    SpectralAR$^\ddagger$~\citep{spectralar}        &\xmark           &4.03                    &-                        &- \\
    TiTok-64$^\dagger$~\citep{titok}                &\xmark           &5.15                    &-                        &- \\
    TiTok-256$^{\dagger*}$~\citep{titok}            &\xmark           &3.84                    &0.542                    &19.92 \\
    \compare{}CVQ$^\dagger$                         &\compare{}\xmark &\compare{}2.60          &\compare{}0.565          &\compare{}20.94 \\
    \midrule
    \multicolumn{5}{l}{\textit{Same enhanced training recipe~\citep{tatitok}}} \\
    TA-TiTok-64~\citep{tatitok}               &\xmark           &2.43                    &-                        &- \\
    TA-TiTok-128~\citep{tatitok}              &\xmark           &1.53                    &-                        &- \\
    TiTok-64~\citep{titok}                    &\xmark           &4.25                    &-                        &- \\
    TiTok-128~\citep{titok}                   &\xmark           &2.63                    &-                        &- \\
    TiTok-256$^*$~\citep{titok}               &\xmark           &1.51                    &0.558                    &20.05 \\
    \compare{}CVQ                             &\compare{}\xmark &\compare{}\textbf{1.29} &\compare{}\textbf{0.569} &\compare{}\textbf{20.97} \\
    \bottomrule
    \end{tabular}}
\end{minipage}
\hfill
\begin{minipage}[ht]{0.49\textwidth}
    \centering
    \captionsetup{type=table}
    \caption{\textbf{FID performance on MJHQ-30K.} We further evaluate text-to-image generation fidelity across different methods on MJHQ-30K.}
    \label{tab:mjhq}
    \tablestyle{5pt}{1.07}
    \resizebox{\linewidth}{!}{
    \begin{tabular}{lcccc}
    \toprule
    Method                        &Type           &1D Token         &Res.          &FID\lowerbetter \\
    \midrule
    SD-XL \citep{sdxl}            &Diff.          &\xmark           &1024          &9.55 \\
    PixArt \citep{pixart}         &Diff.          &\xmark           &512           &6.14 \\
    FlowTok-H~\citep{flowtok}     &Diff.          &\cmark           &256           &7.15 \\
    \noalign{\vspace{1.25pt}}\hdashline\noalign{\vspace{1.25pt}}
    Show-o~\citep{xie2024showo}   &Discrete Diff. &\xmark           &256           &14.99 \\
    TiTok~\citep{titok}           &Mask.          &\cmark           &256           &8.50 \\
    TA-TiTok~\citep{tatitok}      &Mask.          &\cmark           &256           &7.51 \\
    \noalign{\vspace{1.25pt}}\hdashline\noalign{\vspace{1.25pt}}
    STAR~\citep{STAR}             &NSP            &\xmark           &256           &5.67 \\
    SWITTI~\citep{switti}         &NSP            &\xmark           &512           &9.50 \\
    SWITTI~\citep{switti}         &NSP            &\xmark           &1024          &8.10 \\
    \midrule
    LlamaGen~\citep{llamagen}     &NTP            &\xmark           &512           &25.59 \\
    VILA-U~\citep{vilau}          &NTP            &\xmark           &256           &12.81 \\
    Janus~\citep{wu2024janus}     &NTP            &\xmark           &256           &10.10 \\
    UniTok~\citep{unitok}         &NTP            &\xmark           &256           &7.46 \\
    DualToken~\citep{dualtoken}   &NTP            &\xmark           &256           &7.88 \\
    \compare{}CAR                 &\compare{}NTP  &\compare{}\cmark &\compare{}256 &\compare{}\textbf{6.42} \\
    \bottomrule
    \end{tabular}}
\end{minipage}

\vspace{0.2cm}
\section{Conclusion and Future Work}
\label{sec:conclusion}

\textbf{Conclusion~~}
In this work, we introduced CVQ, a simple yet effective vector quantization paradigm that discretizes images along the channel dimension. CVQ achieves high codebook utilization at scale and strong reconstruction fidelity without auxiliary designs. Building upon CVQ, we proposed CAR, an autoregressive image generation framework that shifts the next-token prediction unit from patches to channels. Our results highlight channel-wise tokens as a promising direction for autoregressive image generation and offer new insight into rethinking the fundamental unit of visual tokenization.

\textbf{Limitations~~}
CAR exhibits a distinct failure pattern compared with patch-wise AR. As each channel token spans the entire spatial field, severe channel-prediction errors can affect large image regions, leading to structural distortions or global blurring even when the overall semantics are preserved. In contrast, patch-wise AR failures more often manifest as semantic omissions or localized distortions.

\textbf{Future Work~~}
Despite the encouraging results, several important directions remain for future work.
First, CVQ can be combined with recent advances in VQ, such as SimVQ~\citep{simvq} and IBQ~\citep{ibq}, to further improve representational capacity.
Second, the autoregressive formulation of CAR makes it a natural fit for unified vision models that jointly perform visual understanding and generation within a single architecture.
Finally, extending the channel-wise quantization paradigm to the temporal dimension represents a promising direction for learning efficient and compact video representations.

\section*{Acknowledgments}
This work received computational support from SII and JD.COM and was partially supported by the Zhejiang Leading Innovative and Entrepreneurial Team Introduction Program (2024R01007). The authors declare no competing interests.

\bibliography{neurips_2026}
\bibliographystyle{neurips_2026}

\newpage
\appendix

\section{More t-SNE Visualizations of Embedding Distributions.}
\label{more_viz}

\begin{figure*}[ht]
  \centering
  \includegraphics[width=\textwidth]{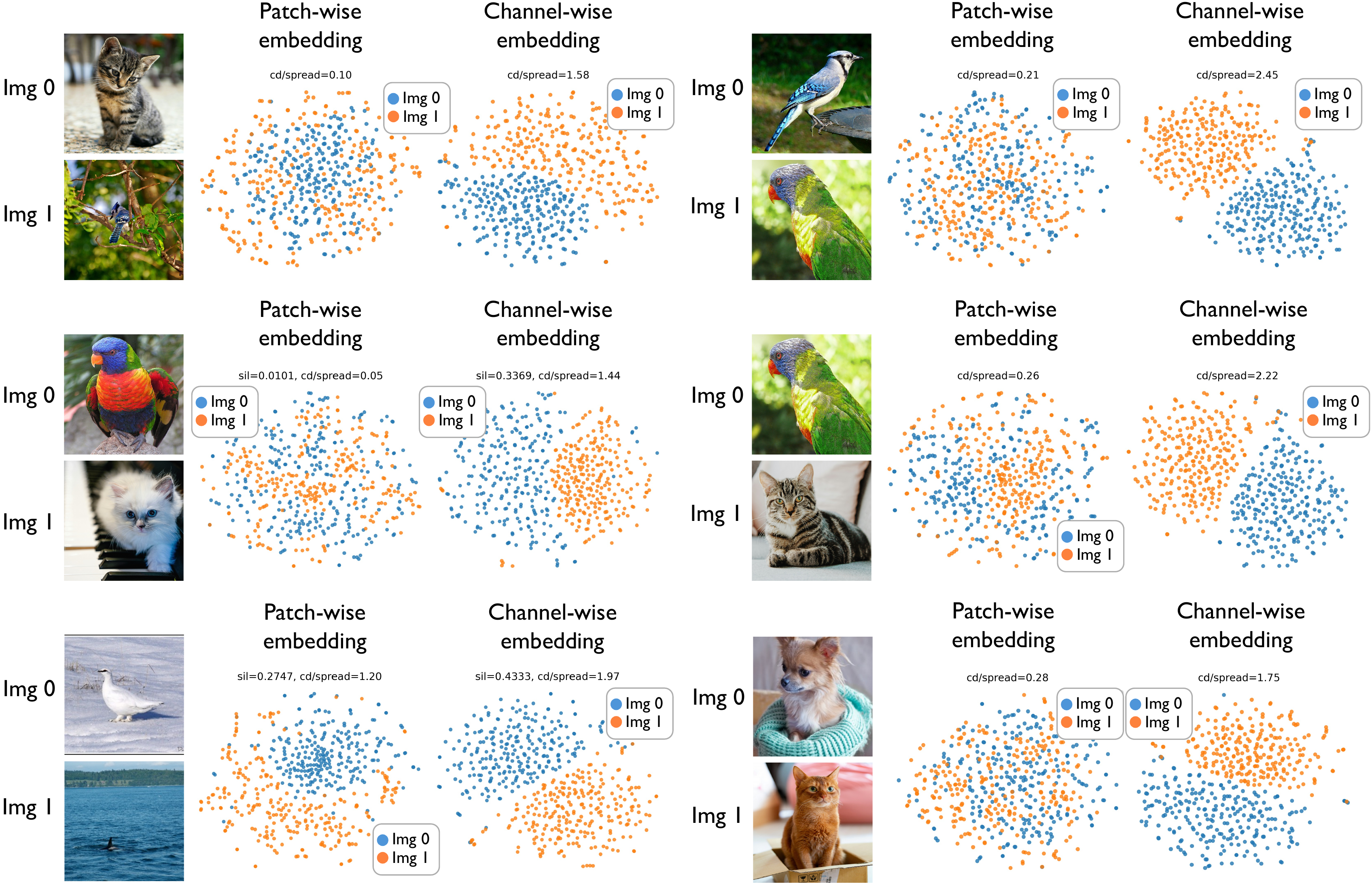}
  \caption{
    More visualizations of feature map channels
  }
  \label{fig:more_viz}
  \vspace{-5pt}
\end{figure*}

\section{Details and Analysis of the Nested Channel Dropout Strategy}
\label{appendix_channel_dropout}

\subsection{Implementation Details of Nested Channel Dropout}

During the training phase, we introduce a structured dropout mechanism to enforce a coarse-to-fine hierarchy within the latent space. For a given latent representation $\mathbf{Z} \in \mathbb{R}^{h \times w \times c}$, we define a truncated configuration where only the first $c_{\text{keep}}$ channels are retained, while the remaining $c - c_{\text{keep}}$ channels are masked (set to zero). This truncated latent, denoted as $\mathbf{Z}_{\perp c_{\text{keep}}}$, effectively forces the model to compress the most critical visual information into the lower-indexed channels. 

The optimization objective for a specific channel configuration $c_{\text{keep}}$ is formulated as follows:
\begin{equation}
    \mathcal{L}_{\text{nested}}(c_{\text{keep}}) = \mathcal{L}_{\text{recon}}(\mathbf{Z}_{\perp c_{\text{keep}}}) + \mathcal{L}_{\text{quant}}(\mathbf{Z}_{\perp c_{\text{keep}}}) + \mathcal{L}_{\text{lpips}}(\mathbf{Z}_{\perp c_{\text{keep}}}) + \lambda_{\text{GAN}}(c_{\text{keep}})\mathcal{L}_{\text{GAN}}(\mathbf{Z}_{\perp c_{\text{keep}}}),
\end{equation}
where $\mathcal{L}_{\text{recon}}$ and $\mathcal{L}_{\text{lpips}}$ denote the standard reconstruction and perceptual losses, respectively. The quantization loss $\mathcal{L}_{\text{quant}}$ is computed exclusively over the active $c_{\text{keep}}$ channels to ensure valid codebook mapping at reduced dimensions.

A key challenge in training nested representations is that the discriminator often suffers from instability when operating on extremely sparse latent features, which can lead to unnatural artifacts in reconstructions using few channels. To mitigate this, we introduce an adaptive GAN weight $\lambda_{\text{GAN}}$ governed by a sigmoid function:
\begin{equation}
    \lambda_{\text{GAN}}(c_{\text{keep}}) = \frac{\lambda_{0}}{1 + e^{-\eta (c_{\text{keep}} - \frac{c}{2})}},
\end{equation}
where $\eta = 0.05$ controls the transition smoothness, and $\lambda_{0} = 1$ is the base GAN weight. This formulation ensures that the adversarial influence is gradually introduced as the channel capacity increases, preserving stable convergence for sparse configurations.

To balance hierarchical representation learning with overall reconstruction fidelity, 
we adopt a stochastic training strategy. In each iteration, nested channel dropout 
is performed with probability $\alpha$ (the dropout ratio), where the number of 
active channels $c_{\text{keep}}$ is sampled from a discrete uniform distribution 
$\mathcal{U}(1, c)$. Conversely, with probability $1-\alpha$, the model undergoes 
standard training using the full channel configuration. The resulting hybrid 
optimization objective is formulated as:
\begin{equation}
\label{lsum}
    \mathcal{L}_{\text{sum}} = \alpha \cdot \mathbb{E}_{c_{\text{keep}} \sim \mathcal{U}(1, c)}[\mathcal{L}_{\text{nested}}(c_{\text{keep}})] + (1 - \alpha) \mathcal{L}_{\text{total}},
\end{equation}
where $\mathcal{L}_{\text{total}}$ represents the loss under the complete channel 
configuration ($c_{\text{keep}} = c$).

\subsection{Analysis of Nested Channel Dropout}

\citep{rippel2014learning} demonstrates that a nested dropout strategy, which stochastically removes nested sets of hidden units, enforces an ordered representation where importance decreases with the dimension index. For semi-linear autoencoders (comprising a linear or sigmoidal encoder with a linear decoder) under $L_2$ loss, they establish that: (i) its optimal solutions are a subset of standard autoencoder optima, preserving global performance (Theorem 1); (ii) the weight matrices must be commutative in their truncation and inversion (Theorem 2); and (iii) with an orthonormality constraint ($\Gamma^{T}\Gamma=\mathbb{I}_{K}$), the model uniquely recovers the PCA solution (Theorem 3).

However, these theoretical guarantees necessitate stringent conditions: the architecture must be semi-linear and optimized via $L_2$ reconstruction loss. While the proposed unit sweeping technique fixes converged units to mitigate gradient decay at higher indices, its iterative nature remains computationally expensive for modern deep learning.

Nevertheless, empirical evidence \citep{bachmann2025flextok,semanticist} suggests that even with deeper architectures, complex loss functions, and advanced frameworks like diffusion models, randomly dropping tokens in a nested manner produces ordered representations that benefit AR generation without significant loss in expressivity. As shown in Table~\ref{tab:dropout_ratio}, reconstruction quality is preserved, indicating no degradation in representational capacity. Additionally, generative performance at 25\% and 50\% dropout exceeds the 0\% baseline, suggesting that structured ordering facilitates AR generation, with similar results at 25\% and 50\% indicating that a moderate degree of nested order is sufficient.

\belowrulesep=0pt\aboverulesep=0pt

\begin{table}[t]
    \centering
    \captionsetup{type=table}
    \caption{
      \textbf{Ablation on nested channel dropout ratio.} We evaluate the impact of varying the dropout ratio $\alpha$ on reconstruction fidelity and downstream text-to-image generation. While reconstruction performance remain stable across different ratios, generation performance improves significantly with the introduction of channel dropout, with comparable gains observed at $\alpha=25\%$ and $\alpha=50\%$.
    }
    \vspace{2mm}
    \label{tab:dropout_ratio}
    \tablestyle{4pt}{1.18}
    \resizebox{0.6\linewidth}{!}{
    \begin{tabular}{c|ccc|cc}
    \toprule
    \multirow{2}{*}{$\alpha$} &\multicolumn{3}{c}{Recon.}                       &\multicolumn{2}{c}{Generation} \\ 
    \cmidrule{2-6}                 &rFID\lowerbetter &SSIM\higherbetter  &PSNR\higherbetter  &Geneval\higherbetter  &DPG\higherbetter \\
    \midrule
     0\%                           &2.60             &0.565              &20.94              &0.62                  &72.76 \\
     25\%                          &2.62             &0.568              &21.01                &0.74~\gain{(+0.12)}   &82.14~\gain{(+9.38)} \\
     50\%                          &2.61             &0.562              &20.88              &0.75~\gain{(+0.13)}                   &81.96~\gain{(+9.20)} \\
    \bottomrule
    \end{tabular}}
\end{table}

\textbf{Dropout Ratio}~~Table~\ref{tab:dropout_ratio} illustrates the influence of the channel dropout ratio on both reconstruction and generation performance. We observe that reconstruction metrics remain stable across the tested range, with no significant degradation in rFID, SSIM, or PSNR as the dropout ratio increases from 0\% to 50\%. However, the generation capability is substantially enhanced by the introduction of dropout; specifically, the GenEval and DPG scores show a marked improvement when moving from 0\% to 25\%, increasing from 0.62 to 0.74 and 72.76 to 82.14, respectively. Further increasing the dropout ratio to 50\% results in comparable generation performance to the 25\% setting, suggesting that a moderate dropout ratio is sufficient to optimize the model's generative robustness without sacrificing its reconstruction fidelity.

\section{Progressive Channel Analysis~~}
\label{appendix_channel_progression}

\begin{table}[ht]
    \centering
    \caption{\textbf{Reconstruction with progressively added channels.}
    We keep only the first $n$ channels before decoding.
    Early channels rapidly recover global appearance and semantic structure, while later channels mainly refine local structure and texture.}
    \vspace{2mm}
    \label{tab:add_channels}
    \tablestyle{5pt}{1.08}
    \resizebox{0.42\linewidth}{!}{
    \begin{tabular}{cccc}
    \toprule
    \# Channels & rFID\lowerbetter & SSIM\higherbetter & PSNR\higherbetter \\
    \midrule
    32  & 30.24 & 0.410 & 16.96 \\
    64  & 11.84 & 0.472 & 18.38 \\
    96  & 6.91  & 0.506 & 19.21 \\
    128 & 4.05  & 0.524 & 19.82 \\
    160 & 3.46  & 0.533 & 20.17 \\
    192 & 3.03  & 0.539 & 20.39 \\
    224 & 2.88  & 0.542 & 20.55 \\
    256 & \textbf{2.63} & \textbf{0.568} & \textbf{21.01} \\
    \bottomrule
    \end{tabular}}
\end{table}

Finally, we analyze how reconstruction changes as more channels are revealed to the decoder.
We progressively keep the first $n$ channels and mask the remaining channels before decoding.
Table~\ref{tab:add_channels} shows a clear coarse-to-fine trend.
From 32 to 128 channels, rFID, SSIM, and PSNR all improve rapidly, indicating that early channels capture global appearance, object identity, and coarse structure.
After 128 channels, rFID improvement slows down, while SSIM and PSNR continue to increase steadily.
This indicates that later channels mainly refine local structure and high-frequency texture, providing quantitative evidence that channel-wise decomposition naturally organizes visual information from global semantics to fine details.

\section{Variable-Resolution Extension}
\label{appendix_variable_resolution}

In the main experiments, we keep a fixed $256 \times 256$ resolution setting to enable strictly controlled comparison with VQ, yielding codewords with a spatial size of $16 \times 16$.

Here, we show a simple practice that extends CVQ to variable resolutions with lightweight resampling modules before and after quantization: before lookup, a fixed set of learnable queries cross-attend features of arbitrary spatial size ($h \times w$) to a fixed size ($h_0 \times w_0$); after lookup, we dynamically generate $h \times w$ target queries to project the quantized channels back to the decoder's desired spatial size. This plays a role analogous to the pre- and post-quantization projections commonly used in patch-wise VQ tokenizers, while preserving CVQ's channel-token formulation.

For variable-resolution training, we progressively scale input resolution from $256$ to $512$ and $1024$. For patch-wise VQ, the embedding dim is fixed at $256$, yielding token counts of $256$, $1024$, and $4096$, respectively. For CVQ, we also fix the embedding dim to $256$ $(16\times16)$ and control the token budget to the intermediate value of the VQ setting (\ie, $1024$) for a fair comparison.

\begin{table}[ht]
    \centering
    \caption{\textbf{Variable-resolution reconstruction.} CVQ is trained with a fixed codeword size and uses resolution-dependent resampling before and after quantization.}
    \vspace{2mm}
    \label{tab:variable_resolution_recon}
    \tablestyle{3pt}{1.05}
    \resizebox{0.55\linewidth}{!}{
    \begin{tabular}{lcccc}
    \toprule
    Method & Train Res. & Inference Res. & rFID\lowerbetter & PSNR\higherbetter \\
    \midrule
    VQ  & 256      & 256  & 4.84 & 19.93 \\
    CVQ & 256      & 256  & \textbf{0.88} & \textbf{25.02} \\
    VQ  & variable & 256  & 4.97 & 19.56 \\
    CVQ & variable & 256  & \textbf{0.92} & \textbf{24.96} \\
    VQ  & variable & 512  & 2.05 & 22.01 \\
    CVQ & variable & 512  & \textbf{0.96} & \textbf{24.55} \\
    \bottomrule
    \end{tabular}}
\end{table}

\vspace{-4mm}

\begin{table}[ht]
    \centering
    \caption{\textbf{Resolution scaling under matched token budgets.} When CVQ is allocated the same number of tokens as VQ at each resolution, its reconstruction quality also improves with resolution.}
    \vspace{2mm}
    \label{tab:matched_budget_resolution}
    \tablestyle{3pt}{1.05}
    \resizebox{0.72\linewidth}{!}{
    \begin{tabular}{lcccc}
    \toprule
    Method & Token Budget & Train Res. & Inference Res. & rFID\lowerbetter / PSNR\higherbetter \\
    \midrule
    VQ  & 256  & variable & 256  & 4.97 / 19.56 \\
    CVQ & 256  & variable & 256  & \textbf{2.63} / \textbf{21.06} \\
    VQ  & 1024 & variable & 512  & 2.05 / 22.01 \\
    CVQ & 1024 & variable & 512  & \textbf{0.96} / \textbf{24.55} \\
    \bottomrule
    \end{tabular}}
\end{table}

As shown in Table~\ref{tab:variable_resolution_recon}, CVQ generalizes across resolutions and consistently outperforms the corresponding VQ baseline. When the token budget is held fixed, CVQ may slightly trade reconstruction fidelity for efficiency at higher resolutions because each token must cover a larger spatial support. This is a controlled quality-efficiency trade-off rather than an inherent inability to model high-resolution images. As shown in Table~\ref{tab:matched_budget_resolution}, when CVQ is allocated the same token budget as VQ at each resolution, reconstruction quality improves with resolution and remains stronger than VQ.

\textbf{Discussion~~}
CVQ provides a practical advantage for autoregressive generation by decoupling token count from image resolution. While patch-wise methods exhibit quadratic token growth as the latent grid scales, CAR maintains nearly constant inference cost by keeping the channel sequence length fixed. Instead of increasing token count, CVQ encodes spatial resolution within each channel token.

\section{Training Setup and Data Sources}
\label{hyperparameters}

The CVQ tokenizer follows the VQGAN-style encoder-decoder architecture~\citep{esser2021vqgan} and is trained on ImageNet-1K at $256\times256$ resolution. With a downsampling factor of $f=16$, CVQ quantizes each channel of a $16\times16\times c$ latent feature map using a shared codebook, where $c\in\{256,1024\}$. Detailed configurations are provided in Table~\ref{tab:tokenizer_recipe}.

\vspace{-0.1cm}

\begin{table}[ht]
    \centering
    \caption{\textbf{CVQ tokenizer configuration and training recipe.}}
    \label{tab:tokenizer_recipe}
    \tablestyle{6pt}{1.08}
    \begin{tabular}{ll}
        \toprule
        Setting & Value \\
        \midrule
        Dataset & ImageNet-1K \\
        Input resolution & $256\times256$ \\
        CVQ latent shape & $16\times16\times256$ or $16\times16\times1024$ \\
        Codeword dimension & 256 \\
        Codebook size & 16,384 \\
        Discriminator & PatchGAN \\
        Optimizer & Adam ($\beta_1=0.5$, $\beta_2=0.9$) \\
        Learning rate / weight decay & $10^{-4}$ / $10^{-4}$ \\
        Batch size / epochs & 256 / 100 \\
        Training hardware & 8 NVIDIA H20 GPUs \\
        \bottomrule
    \end{tabular}
\end{table}

\vspace{0.1cm}

For CAR tokenizers, nested channel dropout is applied with probability $\alpha=0.25$ to induce coarse-to-fine ordering. The retained prefix length is uniformly sampled from $\{1,\ldots,c\}$, with remaining channels zeroed before decoding and quantization loss computed only over active channels. Further details on the training objective are provided in Appendix~\ref{appendix_channel_dropout}.

CAR is trained in two stages using AdamW ($\beta_1=0.9$, $\beta_2=0.96$, weight decay $10^{-3}$). Stage I trains only the MLP projector and LLM head at a learning rate of $10^{-4}$, keeping the LLM backbone frozen. Stage II jointly optimizes all components at a learning rate of $2\times10^{-5}$. The Training data comprises a filtered ImageNet-21K subset, LAION-Aesthetics-12M, CC12M~\citep{changpinyo2021cc12m}, Megalith-10M\footnote{Re-captioned by Qwen3-VL.}, BLIP-3o-short/long~\citep{blip3o}, and 6M in-house aesthetic images.


\end{document}